\documentclass[conference,a4paper]{IEEEtran}
\IEEEoverridecommandlockouts

\usepackage{cite}
\usepackage{amsmath,amssymb,amsfonts}
\usepackage{algorithmic}
\usepackage{graphicx}
\usepackage{textcomp}
\usepackage{xcolor}
\usepackage{booktabs} 
\usepackage{cleveref}
\usepackage{multirow}
\usepackage{caption}
\usepackage{makecell}
\def\BibTeX{{\rm B\kern-.05em{\sc i\kern-.025em b}\kern-.08em
    T\kern-.1667em\lower.7ex\hbox{E}\kern-.125emX}}
\begin{document}

\title{Text2Sign Diffusion: A Generative Approach for Gloss-Free Sign Language Production
}

\author{\IEEEauthorblockN{Liqian Feng}
\IEEEauthorblockA{\textit{School of Computer Science} \\
\textit{The University of Sydney}\\
\quad Camperdown, NSW, Australia\quad \\
lfen0902@uni.sydney.edu.au}
\and
\IEEEauthorblockN{Lintao Wang}
\IEEEauthorblockA{\textit{School of Computer Science} \\
\textit{The University of Sydney}\\
\quad Camperdown, NSW, Australia\quad \\
lwan3720@uni.sydney.edu.au}
\and
\IEEEauthorblockN{Kun Hu\textsuperscript{*} \thanks{* Corresponding author. }}
\IEEEauthorblockA{\textit{School of Science} \\
\textit{Edith Cowan University}\\
\quad\quad Joondalup, WA, Australia \quad\quad\\
k.hu@ecu.edu.au}
\and
\IEEEauthorblockN{\quad\quad\quad\quad\quad\quad\quad\quad Dehui Kong}
\IEEEauthorblockA{\textit{\quad\quad\quad\quad\quad\quad\quad\quad Faculty of Information Technology} \\
\textit{\quad\quad\quad\quad\quad\quad\quad\quad Beijing University of Technology}\\
\quad\quad\quad\quad\quad\quad\quad\quad Beijing, China \\
\quad\quad\quad\quad\quad\quad\quad\quad kdh@bjut.edu.cn}
\and
\IEEEauthorblockN{Zhiyong Wang}
\IEEEauthorblockA{\textit{School of Computer Science} \\
\textit{The University of Sydney}\\
Camperdown, NSW, Australia \\
zhiyong.wang@sydney.edu.au}
}

\maketitle

\begin{abstract}
Sign language production (SLP) aims to translate spoken language sentences into a sequence of pose frames in a sign language, bridging the communication gap and promoting digital inclusion for deaf and hard-of-hearing communities. Existing methods typically rely on gloss, a symbolic representation of sign language words or phrases that serves as an intermediate step in SLP. 
This limits the flexibility and generalization of SLP, as gloss annotations are often unavailable and language-specific. Therefore, we present a novel diffusion-based generative approach - Text2Sign Diffusion (Text2SignDiff) for gloss-free SLP. Specifically, a gloss-free latent diffusion model is proposed to generate sign language sequences from noisy latent sign codes and spoken text jointly, reducing the potential error accumulation through a non-autoregressive iterative denoising process. We also design a cross-modal signing aligner that learns a shared latent space to bridge visual and textual content in sign and spoken languages. This alignment supports the conditioned diffusion-based process, enabling more accurate and contextually relevant sign language generation without gloss. Extensive experiments on the commonly used PHOENIX14T and How2Sign datasets demonstrate the effectiveness of our method, achieving the state-of-the-art performance.
\end{abstract}


\section{Introduction}

Sign languages are the primary communication medium in hard-of-hearing communities, with about 360 million people worldwide experience disabling hearing loss \cite{davis2019hearing}. As visual languages, they use hand movements and facial expressions to convey meaning \cite{liang2023sign}. However, sign languages have distinct grammar and lexicons that differ from spoken languages, making direct mapping difficult \cite{gong2024llms,saunders2020progressive,camgoz2018neural,Camgoz_2020_CVPR,cui2017recurrent}. Thus, bridging this gap requires cross-modal approaches that integrate relevant information from multiple modalities \cite{chen2024locselect}.

\begin{figure}[t]
\begin{center}
  \includegraphics[width=\linewidth]{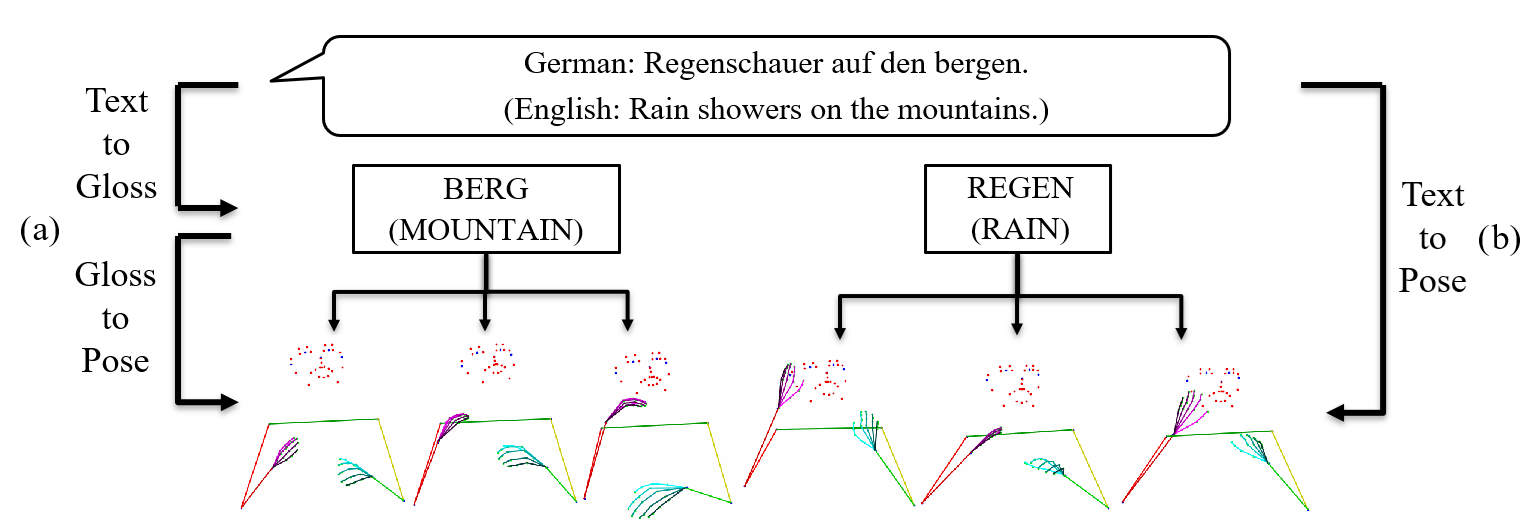}
\end{center}
    \caption{Illustration of Sign Language Production with a sample of German sign language: (a) a two-step approach as text to gloss to pose with intermediate glosses representation; and (b) an one-step gloss-free approach as text to pose directly.}
    \label{fig:1}
\end{figure}

Sign language production (SLP) \cite{Baltatzis_2024_CVPR, Xie_2024_WACV, ma2024attentional} seeks to bridge the gap between spoken and sign languages by translating spoken language into sign pose sequences. To facilitate alignment with spoken language, linguists define glosses, a representation of sign poses, as minimal lexical items in sign languages \cite{huang2021towards}. These representations are widely used in existing SLP works \cite{Arkushin_2023_CVPR, stoll2020text2sign} as intermediates to enhance semantic accuracy, as shown in \Cref{fig:1} (a). 

However, gloss annotation requires expert knowledge to obtain, and the high cost of annotation leads to a scarcity of gloss data in sign language research \cite{lin2023gloss}. Thus, these two-step methods are generally limited to small datasets and poor generalization. Moreover, glosses introduce an information bottleneck between spoken and sign languages, which leads to under-articulation production and limited expressiveness \cite{Baltatzis_2024_CVPR, saunders2021continuous}. Thus, the gloss-free one-step approaches shown in \Cref{fig:1} (b) are of great interest to explore. Furthermore, removing the reliance on gloss faces two main challenges. Firstly, without intermediate gloss representation, models should directly map visual features to spoken language. This direct mapping is inherently abstract and lacks explicit linguistic structure, making it difficult for the model to capture meaningful representations directly from keypoint sequences. Secondly, although several existing gloss-free methods \cite{walsh2024data, saunders2020progressive, Saunders_2021_ICCV, Baltatzis_2024_CVPR, saunders2021continuous, stoll2022there} have been proposed, they often rely on word-level information to synthesize sign language, which struggles to capture sentence-level semantics and align with poses accurately, reducing sign language accuracy. 

Therefore, in this study, we propose Text2Sign Diffusion (Text2SignDiff), a latent diffusion model for gloss-free Sign Language Production, to tackle the above-mentioned challenges: 1) weak end-to-end linguistic-visual translation and 2) poor sentence-level semantics alignment.  For the first challenge, we introduce a novel Gloss-Free Latent Diffusion model which generates continuous sign language by denoising noisy latent sign language representation. Specifically, a transformer-based Sign Variational Autoencoder (SignVAE) model is first learned to embed sign pose sequences within both hand and facial features into a representative latent space. Then, a gloss-free latent diffusion model learns a probabilistic mapping between plain spoken language (texts) and a sequence of latent codes directly. The iterative refinement scheme of diffusion model alleviate the error accumulation issue and enrich diversity in generated sign language. 

For the second challenge, to effectively align the latent pose representations and given spoken language semantics, we devise a novel Cross-Modal Aligner model. It consists of two modules, a Pose Aligner for pose-level sign language meaning encoding and a Text Aligner for sentence-level semantics encoding. They are trained to construct a semantically shared latent space between sign language and spoken language with a contrastive learning scheme. The Cross-Modal Aligner then guides the denoising process by computing a cosine-similarity-based cross-modal semantic loss between noisy poses and overall sentences. The additional guidance is adaptively blended with ordinary diffusion reconstruction at each time step, significantly enhancing the semantic alignment between the generated poses and the input text without requiring intermediate glosses. 

Extensive experiments conducted on the publicly accessible sign language datasets PHOENIX14T \cite{forster2014extensions} and How2Sign \cite{duarte2021how2sign} demonstrate the state-of-the-art generation quality of our proposed Text2Sign Diffusion in comparison with both gloss-based and gloss-free methods. 

In summary, our work's key contributions are as follows:
\begin{itemize}
\item We propose Text2SignDiff, a novel diffusion-based method for Sign Language Production without intermediate gloss representation. 
\item We propose a gloss-free latent diffusion model with SignVAE to iteratively refine and denoise noisy poses in the latent space, generating accurate sign language sequences.
\item We introduce a cross-modal aligner model that enhances semantic accuracy during the denoising process by aligning sign poses and spoken language within a contrastively learned shared latent space.  
\item Comprehensive experimental results demonstrate the effectiveness of the proposed Text2SignDiff.
\end{itemize}

\section{Related Work}
\subsection{Gloss-based Sign Language Production}
A common framework for SLP translates a spoken language sentence into a gloss sentence, which is then used to generate the sign pose sequence as motion data \cite{wang2023multi,Mo_2025_ICCV} in a monotonic manner based on the gloss intermediary. 

Progressive Transformer \cite{saunders2020progressive} introduces the first gloss-based transformer model for SLP, demonstrating great capability in generating continuous sign pose sequences from text. It utilizes an innovative auto-regressive manner, which incrementally generates partial outputs by integrating multiple sources of information using a cross-attention transformer. It serves as a common baseline for subsequent research. An improvement to the Progressive Transformer, incorporating non-manual features such as facial expressions, was explored in \cite{Saunders_2021_ICCV}. It introduces a Mixture of Motion Primitives architecture, consisting of several pose frames as the smallest distinctive structural units in sign languages to represent a gloss word. Another improvement to the Progressive Transformer involves the use of mixture density networks for modeling complex distributions more effectively \cite{saunders2021continuous}. Unlike traditional 2D pose generation, this work introduces the first SLP model to translate spoken language into continuous 3D sign pose sequences. On the other hand, for a non-autoregressive method, NAT-EA \cite{huang2021towards} introduces a parallel decoding scheme called External Aligner for sequence alignment between gloss and pose sequences. It employs gloss duration prediction and a length regulator to match the gloss sequence length with the target sign pose sequence. 

Previous gloss-based SLP frameworks face challenges due to their reliance on fully annotated gloss data, creating a significant data bottleneck. To address this, our Text2SignDiff eliminates the need for gloss intermediaries in both training and inference.
\subsection{Gloss-free Sign Language Production}
NSLP \cite{hwang2021non} introduces a non-autoregressive transformer-based model for SLP that bypasses gloss supervision. The model learns a continuous latent space of sign poses using a Gaussian prior and directly maps spoken language inputs into this space, effectively mitigating issues such as regression to the mean. Recent gloss-free SLP work \cite{walsh2024data} introduces a data-driven representation based on a motion codebook learned from 3D pose data. This approach transforms continuous pose generation into discrete sequence generation, where a progressive transformer translates spoken language into codebook tokens, and a sign-stitching method produces continuous sign poses. 
Unlike this word-level approach, our gloss-free Text2SignDiff model focuses on whole-sentence semantic alignment, improving expression accuracy in SLP tasks through contextual learning.
\subsection{Sign Language Recognition and Translation}
For the reverse process of text-to-pose production, sign language recognition (SLR) translates a specific sign to its corresponding meaning, while Sign Language Translation (SLT) converts a sign pose sequence to its spoken language text \cite{Camgoz_2020_CVPR, Chen_2022_CVPR, chen2022two,HU2025130077,Kan_2022_WACV}. 

Basically, the first application of CNN-based deep learning approaches \cite{koller2019weakly} is utilized for word-level Continuous Sign Language Recognition (CSLR). Expanding from word-level CSLR to sentence-level SLT tasks, researchers have utilized Neural Machine Translation networks \cite{camgoz2018neural, ko2019neural}. Furthermore, transformers also perform well in this translation task. The first end-to-end transformer-based model for SLT \cite{Camgoz_2020_CVPR} introduces a multi-task formalization of CSLR and SLT leveraging the supervision of glosses. Recently, gloss-free settings \cite{yin2023gloss,Zhou_2023_ICCV} reduce the training cost of SLT models without gloss annotations. And with the development of Large Language Models (LLMs), Sign Language Translators\cite{gong2024llms} aims to harness the capabilities of off-the-shelf and frozen LLMs to perform SLT. It regularizes the sign videos into a language-like representation and then prompt the LLM to generate text of the spoken language.

In this work, SLT serves as a module to evaluate SLP performance by translating sign pose frames back to spoken language text and comparing the generated text with the ground truth.


\begin{figure*}[t]
  \centering
    \includegraphics[width=\textwidth]{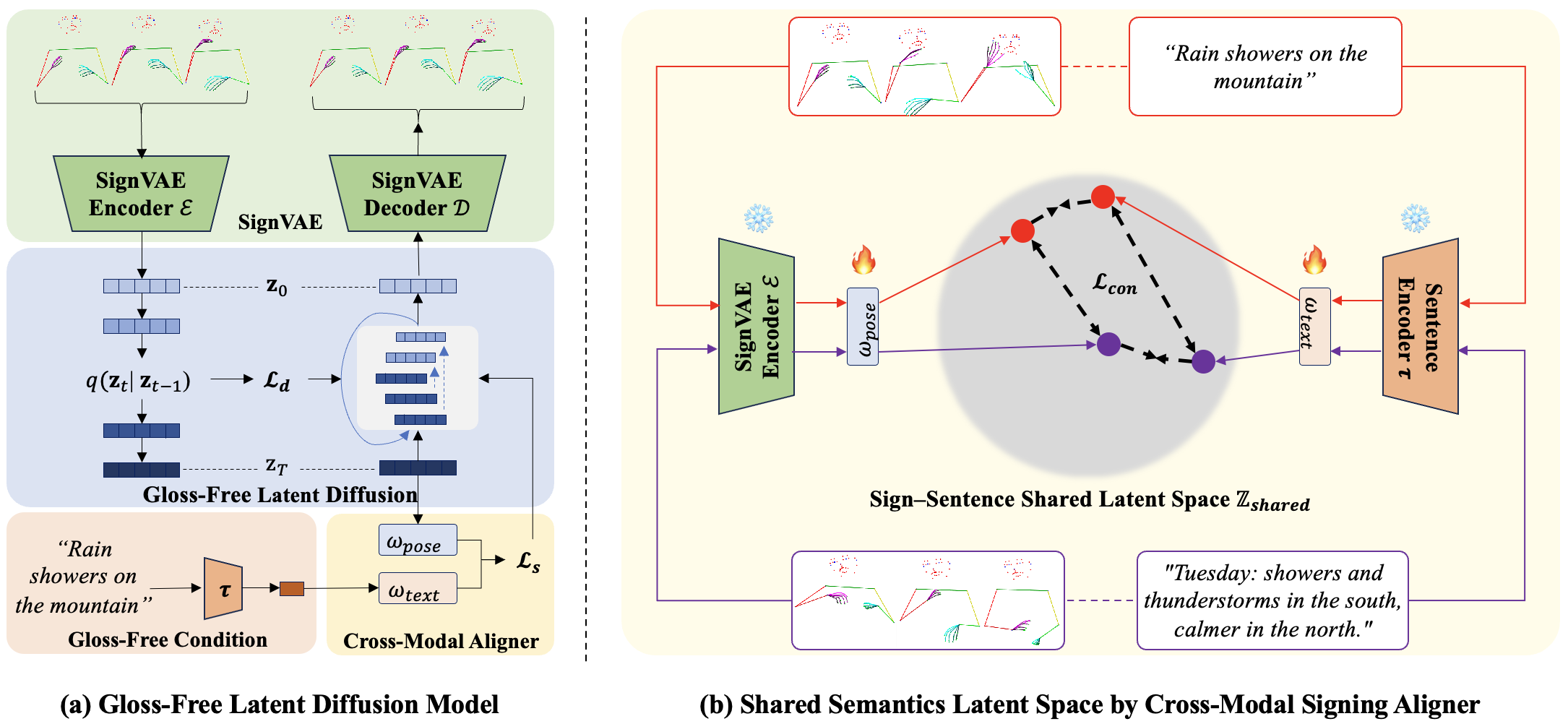}
    \caption{Text2SignDiff Overview: (a) The Gloss-Free Latent Diffusion Model which generates continuous sign language sequences from noisy latent features and gloss-free spoken language conditions; and (b) A shared sign-sentence latent space learned by cross-modal aligner signing with contrastive learning to align sign languages and spoken languages at a sentence-level.}
    \label{fig:arch}
\end{figure*}


\section{Methodology}
The SLP task aims to generate a sign language sentence in poses, denoted as $\mathbf{P} = \{\mathbf{p}_1, \mathbf{p}_2, ..., \mathbf{p}_U\}$ with $U$ frames, from a spoken language sentence in texts, donated as $\mathbf{X} = \{\mathbf{x}_1, \mathbf{x}_2, ..., \mathbf{x}_W\}$ with $W$ words. Particularly, each pose $\mathbf{p}_i$ is defined with $K$ keypoints $\mathbf{p}_i = \{\mathbf{k}^i_1, \mathbf{k}^i_2, ... \mathbf{k}^i_K\}$ and $\mathbf{k}^i_j\in\mathbb{R}^d$ in a ${d}$-dimensional space. 

Our proposed Text2SignDiff, a gloss-free latent diffusion method, consists of three major components: 1) a SignVAE for the latent space representation of sign language pose sequences; 2) a cross-modal aligner to harmonize the pose latent space and the whole sentence semantic latent space; 3) a Denoising Diffusion Probabilistic Model (DDPM) with a deterministic forward diffusion and a reverse denoising process \cite{ho2020denoising}, operating within the latent space defined by SignVAE to synthesize continuous sign language sequences from noises. An overview of the proposed Text2SignDiff is illustrated in \Cref{fig:arch}.

\subsection{Sign Variational Autoencoder}

The Sign Variational AutoEncoder (SignVAE) $\mathcal{V} =\{\mathcal{E, D}\}$ imposes structure on the latent space, providing a regularized, compact representation of sign language pose sequences. It is learned following the standard VAE training process \cite{kingma2013auto,su2025ri}, which consists of an encoder $\mathcal{E}$ and a decoder $\mathcal{D}$. As shown in \Cref{fig:arch} (a), the encoder $\mathcal{E}$ converts a sign pose sequence $\mathbf{P}$ into a latent space as $\mathbf{z}_0 = \mathcal{E}(\mathbf{P}), \mathbf{z}_0 \in\mathbb {R}^{d_{sign}}$, where ${d_{sign}}$ is the dimension of $\mathbf{z}_0$. The decoder $\mathcal{D}$ is to reconstruct the latent representation $\mathbf{z}_0$ into the original sequence of sign poses as $\hat{\mathbf{P}} = \hat{\mathbf{p}}_{1:U} = \mathcal{D}(\mathbf{z}_0)$. We represent the embedded latent distribution as a Gaussian distribution, parameterized by the mean $\mu$ and the variance $\sigma$, which is trained by the sign pose sequence $\mathbf{P}$ reconstruction with the Mean Squared Error (MSE) loss and the Kullback-Leibler (KL) loss as $\mathcal{L}_{vae}$. 
\begin{equation}
\mathcal{L}_{vae} = \|\mathbf{P}-\mathcal{D}(\mathbf{z}_0)\|^2+\operatorname{KL}\left[\mathcal{N}\left(\mu, \sigma^2\right), \mathcal{N}(0,1)\right].
\end{equation}
This allows us to reparameterize the latent space of sign pose sequences. Learning this latent pose representation $\mathbf{z}$ enables compact pose embeddings that ensure both quality and diversity in the generated poses \cite{chen2023executing}. Crucially, this compact space also supports improved semantic alignment with our novel Cross-Modal Signing Aligner. The SignVAE is pre-trained and frozen in the remaining processes, with $\mathbf{z}_0$ as the sign language sequence latent representation for other Text2SignDiff components.

\subsection{Cross-Modal Signing Aligner}

It is a challenge in SLP to establish a sentence-level semantic alignment between spoken language words and corresponding sign poses due to distinct grams and lexicons without gloss intermediates. For better alignment between visual and textual modalities, we devise our new cross-modal signing aligner to learn a joint latent embedding space. As shown in \Cref{fig:arch} (b), there are two encoders: a Pose Aligner $\omega_{pose}$ and a Text Aligner $\omega_{text}$, aligning the pose latent space representations and text sentence embeddings into a shared latent space $\mathbb{Z}_\text{shared}$ in $d_s$ dimension. 

The Pose Aligner takes the latent representations obtained from SignVAE as its input and produces an aligned latent vector $\mathbf{z}_{pose}\in\mathbb{R}^{d_s}$. For the input spoken language in text format, we first obtain the sentence embedding by Sentence-BERT \cite{reimers2019sentence, choi2021evaluation}. Correspondingly, the Text Aligner utilizes the sentence embedding as its input and produces an aligned latent vector $\mathbf{z}_{text}\in\mathbb{R}^{d_s}$ in $\mathbb{Z}_\text{shared}$. 

\begin{equation}
\begin{aligned}
    \mathbf{z}_{pose} &= \omega_{pose}(\mathcal{E}(\mathbf{P})), \\
    \mathbf{z}_{text} &= \omega_{text}(\tau(\mathbf{X})).
\end{aligned}
\end{equation}

To properly train aligners for a shared sign-sentence latent space, we adapt cosine similarity to assess the correspondence between latent features and contrastive-based batch-wise InfoNCE loss. As illustrated in \Cref{fig:arch} (b), given sign-sentence pairs, the contrastive objective $\mathcal{L}_{con}$ aims to maximize (pull) the cosine similarity of the associated pairs and minimize (push) the cosine similarity of the non-related pairs within a batch. Notably, during Pose Aligner and Text Aligner trained by $\mathcal{L}_{con}$, encoder $\mathcal{E}$ is frozen. 

\begin{equation}
\begin{array}{c}
\mathcal{L}_{con} = - \sum\limits_{i} \log \frac{\exp\left( \mathrm{sim}\left(\mathbf{z}_{pose_i}, \mathbf{z}_{text_i}\right) \right)}{\sum\limits_{j} \exp\left( \mathrm{sim}\left(\mathbf{z}_{pose_i}, \mathbf{z}_{text_j}\right)\right) },\\
\mathrm{sim}(\mathbf{z}_{pose}, \mathbf{z}_{text}) = \frac{ \mathbf{z}_{pose}\cdot \mathbf{z}_{text}}{ || \mathbf{z}_{pose}|||| \mathbf{z}_{text}||}.
\end{array}
\end{equation}

\subsection{Gloss-free Latent Diffusion Model}

Our Text2SignDiff follows a conditional latent diffusion model design \cite{ho2020denoising,lu2024autoregressive}, with an additional cross-modal semantic alignment loss, which can gradually transform a Gaussian noise distribution into the target signing data distribution through a Markov process: a forward diffusion process that perturbs sign language latent representation $\mathbf{z}_0$ into noise $\mathbf{z}_T$ in $T$ time steps, and a reverse denoising process that converts noise back into signing data \cite{ho2020denoising}. The forward diffusion process transforms the latent space distribution $\mathbf{z}_0 \sim q(\mathbf{z})$ into a simple prior normal distribution $\mathcal{N}(0,\mathbf{I})$. A sequence of noisy features $\{\mathbf{z}_1, \mathbf{z}_2, ..., \mathbf{z}_T\}$ in $T$ timesteps is created with the transition kernel:
\begin{equation}
    q(\mathbf{z}_t|\mathbf{z}_{t-1}) =  \mathcal{N}(\mathbf{z}_t;(\sqrt{1-\beta_t})\mathbf{z}_{t-1},\beta_t\mathbf{I}),
\end{equation}
where $\beta_t\in(0,1)$ is a series of hyper-parameters for sampling. 
The joint distribution, denoted as $q(\mathbf{z}_1, ...,\mathbf{z}_T|\mathbf{z}_0)$ conditioned on $\mathbf{z}_0$, is factorized as
\begin{equation}
    q(\mathbf{z}_1, ...,\mathbf{z}_T|\mathbf{z}_0) = \prod_{t=1}^{T}q(\mathbf{z}_t|\mathbf{z}_{t-1}).
\end{equation}

The reverse denoising process includes additional whole-sentence semantic meaning $\mathbf{y} = \tau(\mathbf{X})$ condition. The joint distribution $p_\theta(\mathbf{z}_{0:T}|\mathbf{y})$ is defined as a Markov chain, starting with $p(\mathbf{z}_T) = \mathcal{N}(\mathbf{z}_t;0,\mathbf{I})$ and then iteratively sampling from the learnable transition kernel $p_\theta$: 
\begin{equation}
    p_\theta(\mathbf{z}_{0:T}|\mathbf{y}) = p(\mathbf{z}_T)\prod_{t=1}^{T}p_\theta(\mathbf{z}_{t-1}|\mathbf{z}_t,\mathbf{y}).
\end{equation}

The learnable transition kernel $p_\theta(\mathbf{z}_{t-1}|\mathbf{z}_t,\mathbf{y})$ takes the form of:
\begin{equation}
    p_\theta(\mathbf{z}_{t-1}|\mathbf{z}_t,\mathbf{y}) = \mathcal{N}(\mathbf{z}_{t-1};\mu_\theta(\mathbf{z}_t,t,\mathbf{y}),\Sigma_\theta(\mathbf{z}_t,t,\mathbf{y})).
\end{equation}
where $\theta$ denotes model parameters and the mean $\mu_\theta(\mathbf{z}_t,t,\mathbf{y})$ is learned by a U-Net with skip connection and additional sentence semantic condition $\mathbf{y}$ is incorporated by cross-attention \cite{Rombach_2022_CVPR}.

For optimization, the standard DDPM loss function calculates the MSE of $\epsilon$ and $\hat\epsilon_\theta(\mathbf{z}_t,t,\mathbf{y})$
\begin{equation}
    \mathcal{L}_{d} = \mathbb{E}_{\epsilon,t}[||\epsilon-\hat\epsilon_\theta(\mathbf{z}_t,t,\mathbf{y})||^2].
\end{equation}

In addition to traditional DDPM loss, Text2SignDiff utilizes the cross-modal aligner model to provide additional semantic supervision.
\begin{equation}
\begin{aligned}
    \mathbf{z}_{pose} &= \omega_{pose}(\hat{\mathbf{z}}), \\
    \mathbf{z}_{text} &= \omega_{text}(\tau(\mathbf{X})), \\
    \mathcal{L}_{s} &= 1-\frac{\mathbf{z}_{pose} \cdot \mathbf{z}_{text}}{||\mathbf{z}_{pose}||||\mathbf{z}_{text}||}.
\end{aligned}
\end{equation}
The cross-modal semantic alignment loss is combined with the diffusion loss using a time-based factor, $(1-\frac{t}{T})$, which applies stronger semantic supervision in early denoising steps, guiding latent noise toward a semantically accurate direction.  For a certain time step $t$, the loss function $\mathcal{L}$ is defined as: 
\begin{equation}
    \mathcal{L} = \mathcal{L}_{d} + (1-\frac{t}{T})*\mathcal{L}_{s}.
    \label{eq:eq_loss}
\end{equation}

\begin{table*}[t]
  \centering
  \begin{tabular*}{\textwidth}{@{\extracolsep{\fill}}lcccccc}
    \toprule
    Method &  BLEU-4↑ & BLEU-3↑ & BLEU-2↑ & BLEU-1↑ & ROUGE↑ & DTW↓\\
    \midrule
    Ground Truth$^\dag$ & 19.45 & 23.92 & 31.05 & 43.72 & 45.44 & 0.0000\\
    \midrule
    Progressive Transformers \cite{saunders2020progressive} & 4.04 & 5.08 & 7.08 & 11.45 & 13.17 & 0.1910\\
    GEN \cite{tang2022gloss} & 6.20 & 8.09 & 11.53 & 18.71 & 19.79 & 0.1189\\
    DET \cite{viega_2023} & 5.76 & 7.39 & 10.39 & 17.18 & 17.64 & -\\
    NAT-EA \cite{huang2021towards} & 6.66 & 7.99 & 10.45 & 15.12 & 19.43 & 0.1460\\
    G2P-DDM \cite{xie2024g2p} & 7.50 & 9.22 & 11.37 & 16.11 & - & 0.1160\\
    CasDual-Transformer \cite{ma2024attentional}  & \underline{10.00} & \underline{12.47} & \underline{16.73} & 25.14 & 26.12 & - \\
    Data-Driven Representation SLP \cite{walsh2024data} & 9.20 & 11.75 & 16.36 & \underline{27.24} & \underline{27.93} & \underline{0.1050}\\
    \midrule
    \textbf{Text2SignDiff (Ours)} & \textbf{11.23} & \textbf{15.18} & \textbf{20.91} & \textbf{32.65} & \textbf{33.08} & \textbf{0.0997}\\
    \bottomrule
  \end{tabular*}
  \caption{Quantitative evaluation on the PHOENIX14T dataset. ``\dag'' refers to the reconstructed results of the back-translation SLT model on ground truth sign poses.}
  \label{tab:quant1}
\end{table*}

\begin{table*}[t]
  \centering
  \begin{tabular*}{\textwidth}{@{\extracolsep{\fill}}lcccccc}
    \toprule
    Method &  BLEU-4↑ & BLEU-3↑ & BLEU-2↑ & BLEU-1↑ & ROUGE↑ & DTW↓\\
    \midrule
    Ground Truth$^\dag$ & 9.93 & 12.67 & 17.96 & 28.37 & 28.94 & 0.0000\\
    \midrule
    Progressive Transformers \cite{saunders2020progressive} & 2.01 & 3.86 & 7.04 & 13.69 & 13.81 & -\\
    Progressive Transformers w/ MS \cite{Saunders_2021_ICCV} & 2.34 & 3.92 & 7.63 & 13.68 & 13.83 & -\\
    Ham2Pose \cite{Arkushin_2023_CVPR} & 2.93 & 4.07 & 7.31 & 12.38 & 13.29 & -\\
    T2M-GPT \cite{zhang2023generating} & 3.53 & 5.14 & 7.92 & 12.87 & 13.99 & -\\
    MS2SL-T2S \cite{ma2024ms2sl} & \underline{4.26} & \underline{6.84} & \underline{9.17} & \underline{14.67} & \underline{16.38} & -\\
    \midrule
    \textbf{Text2SignDiff (Ours)} & \textbf{5.68} & \textbf{8.28} & \textbf{12.11} & \textbf{17.91} & \textbf{18.33} & \textbf{0.1051}\\
    \bottomrule
  \end{tabular*}
  \caption{Quantitative evaluation on the How2Sign dataset. ``\dag'' refers to the reconstructed results of the back-translation SLT model on ground truth sign poses.}
  \label{tab:quant2}
\end{table*}

\begin{table*}[t]
  \centering
  \begin{tabular*}{\textwidth}{@{\extracolsep{\fill}}lcccc|cccc}
    \toprule
    \multirow{2}{*}{Methods} &
    \multicolumn{4}{c|}{PHOENIX14T} &
    \multicolumn{4}{c}{How2Sign} \\
    \cmidrule(lr){2-9}
    &BLEU-4↑&BLEU-1↑&ROUGE↑&DTW↓
    &BLEU-4↑&BLEU-1↑&ROUGE↑&DTW↓\\
    \midrule
    \textbf{Text2SignDiff (Ours)} & \textbf{11.23} & \textbf{32.65} & \textbf{33.08} & \textbf{0.0997} & \textbf{5.68} & \textbf{17.91} & \textbf{18.33} & \textbf{0.1051}\\
    w/o pose-text aligner pre-training & 8.32 & 21.25 & 20.91 & 0.1202 & 3.29 & 13.09 & 13.21 & 0.1250\\
    w/o diffusion time-factor & 9.91 & 28.83 & 29.09 & 0.1031 & 4.97 & 14.79 & 15.28 & 0.1105\\
    diffusion w/o pose-text aligner & 6.35 & 15.83 & 15.13 & 0.1434 & 3.17 & 12.09 & 12.37 & 0.1308\\
    \bottomrule
  \end{tabular*}
  \caption{Ablation study on the PHOENIX14T and How2Sign dataset.}
  \label{tab:ablation}
\end{table*}

\section{Experiments}
\subsection{Experimental Setting}
\noindent\textbf{Dataset.} We evaluate our model on two publicly available benchmark datasets. PHOENIX14T \cite{camgoz2018neural}, derived from German weather broadcasts, contains 8,257 sign pose clips with a vocabulary of 2,887 German words, and How2Sign \cite{duarte2021how2sign}, an American Sign Language dataset, comprises over 34,000 clips across various instructional scenarios, covering more than 16,000 English words. These datasets serve as primary benchmarks for evaluating sign language translation and production models in recent research.

\noindent\textbf{Implementation Details.} For sign language pose features, each pose $\mathbf{p}_i$ consists of 79 keypoints: 21 for each hand, 11 for the body, and 26 for the face. We set a length limit $U$ of 256 frames for Text2SignDiff's training and inference sequences. Specially, we pad the end with a unique End-of-Sign frame. To generate variable-length sequences within the $U$-frame limit, we detect the End-of-Sign frame in the output and truncate the sequence at that point, removing any blank frames from the End-of-Sign frame onward.

For the SignVAE model, pre-trained for 1000 epochs, creates a 20-dimensional latent space by encoding pose clips, which is used by the Pose Aligner and latent diffusion module. The text features are extracted by distiluse-base-multilingual-cased-v2 \cite{reimers2019sentence, choi2021evaluation} Sentence-BERT model, mapping every text sentence to a 512-dimensional vector. The Pose Aligner and Text Aligner then are pre-trained for 1000 epochs to align the 20-dimensional pose latent space and 512-dimensional sentence embedding space into a new shared 512-dimensional latent space $\mathbb{Z}_\text{shared}$. In training the gloss-free latent diffusion model, we utilize frozen SignVAE and cross-modal aligner components. For the diffusion process, we follow the DDPM setup with a linear noise schedule. The forward process is discretized into 1000 timesteps, with noise variance linearly spaced from 1e-4 to 0.02. We train the model using a batch size of 256 for 1000 epochs by $\mathcal{L}$ in \Cref{eq:eq_loss}. The Adam optimizer is employed with a learning rate of 1e-4. All experiments are conducted using PyTorch on an NVIDIA RTX A6000 GPU (48GB).

\noindent\textbf{Evaluation Metrics.}
In typical SLP tasks, the back-translation evaluation metric is used to assess generation performance \cite{walsh2024data}. Specifically, it utilizes a pre-trained SLT model \cite{Camgoz_2020_CVPR} to translate the produced sign pose sequences back to spoken language. BLEU-n (from 1 to 4) and ROUGE are then calculated between back-translated text and ground truth. They are widely used metrics to assess the quality of machine-translated text against a reference text \cite{papineni2002bleu}. 
To assess the accuracy of generated sign poses, we also use mean joint error with Dynamic Time Warping (DTW) \cite{berndt1994using} to compute the average distance (0 to 1) between generated sign pose features and ground truth, as done in prior SLP research \cite{xie2024g2p, walsh2024data}. Unlike BLEU, which evaluates text correspondence, DTW objectively measures the spatial alignment of each keypoint's coordinates.


\begin{table*}[t]
  \centering
  \begin{tabular*}{\textwidth}{@{\extracolsep{\fill}}ll}
    \toprule
    \multicolumn{2}{c}{Input Text: Gelegentlich schneit es leicht richtung nordosten kann es auch noch kräftiger schneien.}\\
    \multicolumn{2}{c}{(English: Occasionally it snows lightly, towards the northeast it can snow even heavier.)}\\
    ~ & \multirow{5}{*}{\includegraphics[width=0.82\textwidth]{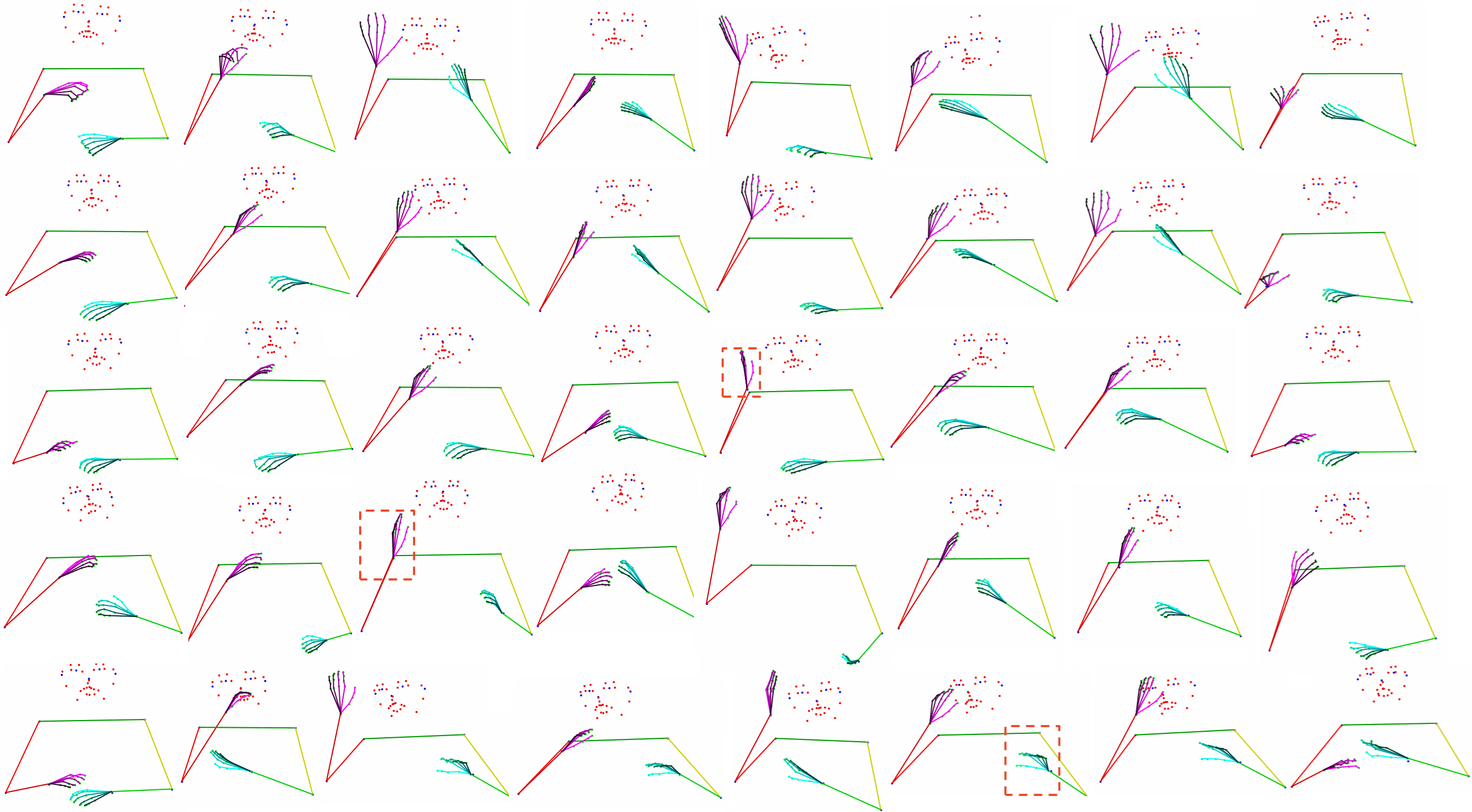}}\\
    \rule{0pt}{15pt}Ground Truth & ~\\
    \rule{0pt}{42pt}\textbf{T2SDiff (Ours)} & \\
    \rule{0pt}{42pt}\makecell[l]{w/o Pose-Text\\ Aligner Pre-training} & \\
    \rule{0pt}{42pt}\makecell[l]{w/o Diffusion\\ Time-Factor} & \\
    \rule{0pt}{40pt}\makecell[l]{Diffusion\\ w/o Pose-Text Aligner} & \\
    ~&~\\
    \\
    \midrule
    \multicolumn{2}{c}{Input Text: Wash your face with fresh water.}\\
    ~ & \multirow{5}{*}{\includegraphics[width=0.82\textwidth]{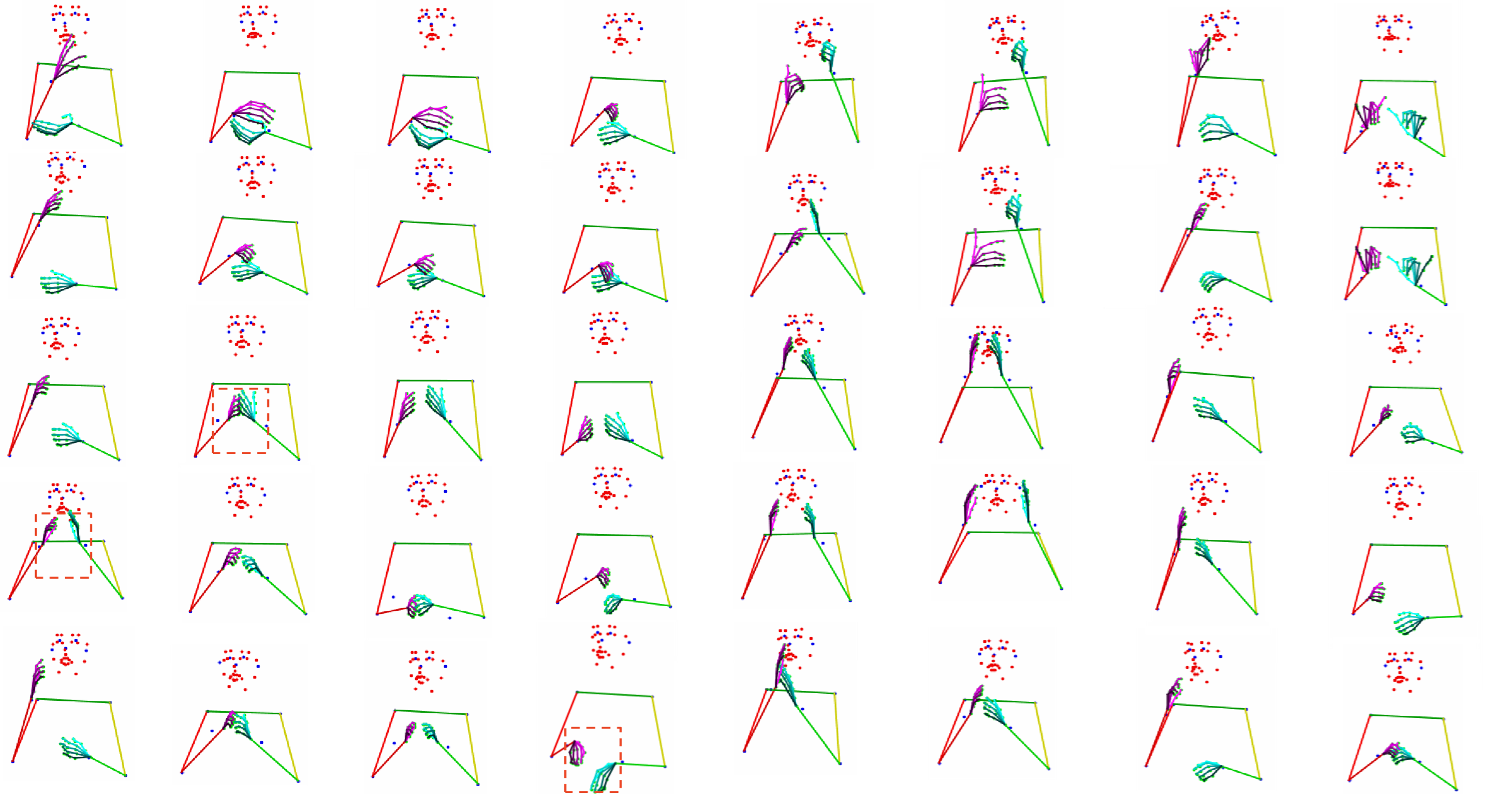}}\\
    \rule{0pt}{15pt}Ground Truth & ~\\
    \rule{0pt}{38pt}\textbf{T2SDiff (Ours)} & \\
    \rule{0pt}{38pt}\makecell[l]{w/o Pose-Text\\ Aligner Pre-training} & \\
    \rule{0pt}{38pt}\makecell[l]{w/o Diffusion\\ Time-Factor} & \\
    \rule{0pt}{40pt}\makecell[l]{Diffusion\\ w/o Pose-Text Aligner} & \\
    ~&~\\
    \\
    \bottomrule
  \end{tabular*}
  \caption{Qualitative examples. We show input spoken language sentences and the corresponding generated sign language poses. Errors or misalignments in the ablation study of Text2SignDiff are highlighted with red boxes.}
  \label{tab:quali}
\end{table*}

\subsection{Comparison with State-of-the-Art}

We compare our Text2SignDiff with several existing SLP methods, including both gloss-based \cite{saunders2020progressive, tang2022gloss, viega_2023, huang2021towards, xie2024g2p} and gloss-free \cite{ma2024attentional, walsh2024data, zhang2023generating, ma2024ms2sl} methods, using BLEU-n, ROUGE, and DTW as evaluation metrics. The quantitative results on PHOENIX14T and How2Sign datasets are presented in \Cref{tab:quant1} and \Cref{tab:quant2}. Our model outperforms prior methods, achieving the highest BLEU-n and ROUGE score, and the lowest DTW score. Compared to gloss-based methods, Text2SignDiff eliminates the dependency on manual gloss annotations while achieving better semantic and temporal alignment. Furthermore, our model outperforms recent gloss-free approaches by better aligning sentence-level text semantics with sign pose representations, benefiting from our shared latent space and cross-modal aligner module. These results surpass both gloss-based and gloss-free baselines indicating superior generation quality and robustness, and demonstrate the model’s ability to align semantic and sign pose modality.

\subsection{Ablation Study}
In this section, we conduct ablation studies of the proposed modules to validate their effectiveness. As discussed, the Text2SignDiff consists of three major components and trained in sequential way. The SignVAE is first pre-trained to embed pose sequences into latent space, which will be frozen in the following steps. Then, the cross-modal aligner model is pre-trained in a contrastive manner to establish a shared latent space between pose and text. Finally, the latent diffusion is trained with cross-modality model to synthesize sign languages without gloss intermediates. 

We first ablate the pre-training of cross-modal aligner model, training it only alongside the gloss-free latent diffusion. As shown in \Cref{tab:ablation}, all metrics show significant degradation, as the cross-modal aligner fails to provide meaningful semantic guidance during the initial stages of the diffusion process. This highlights the importance of establishing a shared latent space between sign and sentence for improved semantic alignment. 

Additionally, we perform an ablation study on the diffusion time factor when integrating the semantic loss function with the diffusion loss, as described in \Cref{eq:eq_loss}. Similarly, all metrics degrade, although at a smaller scale compared to the diffusion stage within the factor. This time factor incorporates the semantic loss into the diffusion process in a time-discounted manner, providing stronger semantic supervision during the initial stages of diffusion when the latent codes are highly noisy. The results demonstrate that this discounting factor is essential for guiding the latent codes towards accurate semantics and generating high-quality sign language outputs.

Finally, we ablate the cross-modal aligner during latent diffusion model training, directly applying a conditional DDPM for SLP. It yields suboptimal performance compared to the diffusion process within the pre-trained cross-modal aligner model. This result underscores the importance of our design choice of the cross-modal aligner, as it enhances semantic alignment with noisy latent codes during diffusion, supporting more accurate alignment.

\subsection{Qualitative Evaluation}
\Cref{tab:quali} shows two examples from spoken language sentences to sign language pose sentences with PHOENIX14T and How2Sign datasets. We selected a number of frames from the sign pose sequence as a showcase and compared them with the ground truth. Specifically, we focus on the details of the movements of the generated hand poses. 

The first example translates the spoken language sentence: "\textit{Gelegentlich schneit es leicht richtung nordosten kann es auch noch kräftiger schneien}" (in English ``\textit{Occasionally it snows lightly, towards the northeast it can snow even heavier}''), while both hands waving up and down convey `\textit{snow}' and a hand points to the \textit{northeast} in sign language. Our model clearly expresses the motion and movement details of the poses. A similar situation can also be observed in the second example for translation ``\textit{Wash your face with fresh water}'', in which a hand moving around the face means `\textit{Wash your face}'.

Overall, the qualitative results demonstrate that our Text2SignDiff model avoids error accumulation and, without gloss annotations, effectively learns the whole-sentence meaning of the text to produce expressive sign pose sequences.

\section{Limitation \& Future Work}
One major limitation of this study is the length of the produced pose frames. Currently, our method can generate a sign pose sequence of up to 256 frames per input sentence. While this may be sufficient for texts divided sentence by sentence in the dataset we used, the frame limitation constrains its performance in more complex input scenarios. Additionally, a higher frame rate (frames per second) is required for smoother motion presentation in various applications. Thus, exploring diffusion methods to enhance its capability for generating longer-duration temporal sequences in sign languages could further benefit the community.

Furthermore, although our method can produce sign language sentences within seconds, it still falls short of meeting real-time application requirements. Enhancing the approach with a more efficient generation design using diffusion methods and an optimized embedding procedure could broaden its applicability across a wider range of scenarios.

\section{Conclusion}
The SLP model bridges the communication gap between the hearing and hard-of-hearing communities and can potentially increase their inclusiveness in the social environment. In this work, we present a novel diffusion model Text2SignDiff for Gloss-free Sign Language Production. Compared with previous auto-regressive methods, our Text2SignDiff runs in a non-autoregressive manner and can overcome the error accumulation. Furthermore, without gloss intermedia, Text2SignDiff utilizes the whole-sentence semantic meaning as supervision and aligns the sign pose representation and text sentence embedding, which is not limited by the performance of gloss annotating. Extensive experiments demonstrate the state-of-the-art back-translation and DTW performance.

\section*{Acknowledgments}
This work was partially supported by the Edith Cowan University Science Early Career and New Staff Grant Scheme.

\bibliographystyle{IEEEtran}
\bibliography{IEEEabrv,IEEEfull}

\end{document}